\documentclass{article}

\usepackage[preprint]{neurips_2026}

\usepackage[utf8]{inputenc}
\usepackage[T1]{fontenc}
\usepackage{hyperref}
\usepackage{url}
\usepackage{booktabs}
\usepackage{amsfonts}
\usepackage{amsmath}
\usepackage{nicefrac}
\usepackage{microtype}
\usepackage{xcolor}
\usepackage{graphicx}
\usepackage{listings}
\usepackage{enumitem}
\usepackage{caption}
\usepackage{subcaption}
\usepackage{float}
\usepackage{epigraph}

\definecolor{codegreen}{rgb}{0.0, 0.5, 0.0}
\definecolor{codegray}{rgb}{0.5, 0.5, 0.5}
\definecolor{codepurple}{rgb}{0.58, 0, 0.82}

\title{The Discrete Charm of the MLP:\\
  Binary Routing of Continuous Signals in Transformer Feed-Forward Layers}

\author{
  Peter Balogh \\
  \texttt{palexanderbalogh@gmail.com}
}

\newcommand{\codeurl}{\url{https://github.com/pbalogh/discrete-charm-mlp}}

\begin{document}

\maketitle

\begin{quote}
\emph{``It just happened that no one else was familiar with both fields at the same time.''}
\hfill ---Claude Shannon
\end{quote}

\begin{abstract}
We show that MLP layers in transformer language models perform \emph{binary routing of
continuous signals}: the decision of whether a token needs nonlinear processing is
well-captured by binary neuron activations, even though the signals being routed are
continuous.
In GPT-2 Small (124M parameters), we find that specific neurons implement a consensus
architecture---seven ``default-ON'' neurons and one exception handler (N2123 in Layer~11)
that are 93--98\% mutually exclusive---creating a binary routing switch.
A cross-layer analysis reveals a developmental arc: early layers (L1--3) use single
gateway neurons to route exceptions without consensus quorums; middle layers (L4--6)
show diffuse processing with neither gateway nor consensus; and late layers (L7--11)
crystallize full consensus/exception architectures with increasing quorum size
(1$\to$3$\to$7 consensus neurons). Causal validation confirms the routing is
functional: removing the MLP at consensus breakdown costs 43.3\% perplexity,
while at full consensus removing it costs only 10.1\%---exceeding a $4\times$ difference. Comparing binary vs.\ continuous features for
the routing decision confirms that binarization loses essentially no information (79.2\%
vs.\ 78.8\% accuracy), while continuous activations carry additional magnitude information
($R^2 = 0.36$ vs.\ $0.22$). This binary routing structure \emph{explains} why smooth
polynomial approximation fails: cross-validated polynomial fits
(degrees 2--7) never exceed $R^2 = 0.06$ for highly nonlinear layers.
We propose that the well-established piecewise-affine characterization of deep networks
\citep{balestriero2018spline} can be complemented by a \emph{routing} characterization:
along the natural data manifold, the piecewise boundaries implement binary decisions about
which tokens need nonlinear processing, routing continuous signals through qualitatively
different computational paths.\footnote{Code and data available at \codeurl}
\end{abstract}

\section{Introduction}

\subsection{The Smooth Function Framing}

The standard view of transformer MLPs treats them as function approximators. The residual
stream presents what looks like a function-approximation problem: the MLP receives a
768-dimensional input and must produce a 768-dimensional output, pointwise, for each token
position. It is natural to frame this as curve-fitting: the MLP uses its 3072 piecewise-linear
(GELU) neurons to approximate whatever smooth function maps inputs to outputs along the data
manifold. The universal approximation theorem assures us this is possible; the question becomes
how efficiently the network achieves it.

The most elegant formalization of this view is \citet{balestriero2018spline}, who prove
rigorously that deep networks with piecewise-linear activations compute continuous
piecewise-affine spline functions---a result that connects neural networks to
classical approximation theory. Under the spline view, the network partitions its input space
into polytopes and fits an affine (degree-1) function within each. This characterization is
mathematically precise and correct: every forward pass \emph{does} trace a path through a
piecewise-affine partition.

We take this characterization as our starting point and ask a complementary question:
what kind of \emph{computation} does the piecewise structure implement?
The spline framework tells us the MLP partitions input space into regions with different
affine maps. But it does not tell us whether those regions reflect smooth variation along the
data manifold---the MLP approximating a continuous function with increasing local
resolution---or whether they reflect discrete \emph{decisions}: binary conditions that
route tokens to qualitatively different processing.

Consider a concrete analogy. A 2D shape---say, the infinity symbol ($\infty$)---projected
onto a 1D number line collapses points near the crossing: distinct locations on the upper and
lower loops map to the same coordinate. The MLP must resolve this ambiguity. Under the smooth
function framing, we expect it to do so by tracing the curve through the
intersection---separate branches, each locally smooth, recoverable by the right clustering
or polynomial fit. We find instead that it does so by \emph{flipping a switch}: upper or
lower, a binary decision.

The MLP's $4\times$ width expansion temporarily provides more dimensions, and the nonlinear
activation creates the region boundaries---but the character of those boundaries turns out to
be discrete routing, not smooth approximation.

\subsection{The Kolmogorov Connection}

The Kolmogorov representation theorem guarantees that any continuous multivariate function can
be decomposed into compositions of univariate functions and addition. An MLP literally implements
this: linear combination $\to$ pointwise nonlinearity $\to$ repeat. But the theorem says nothing
about the \emph{character} of the univariate functions needed. If the underlying computation is
low-degree polynomial, the MLP is wastefully approximating it with thousands of linear pieces.
If it is high-degree or non-polynomial, the piecewise approximation may be the most efficient
representation available. And if the computation is fundamentally discrete---a set of binary
routing decisions---then the piecewise structure is not approximating anything at all; it is
implementing a switching network.

We set out to determine which case holds in practice by looking for smooth structure in the
nonlinear component, and simultaneously looking for discrete structure in the neuron
activations. The smooth structure is absent; the discrete structure is present and interpretable.

\subsection{An Analogy: Shannon's Switch}

In 1937, Claude Shannon showed that relay switches---continuous electromechanical devices---could
implement Boolean algebra: each switch routes current or blocks it, and any arrangement of
switches implements a logical expression \citep{shannon1938symbolic}. Before Shannon, engineers
analyzed relay circuits using the continuous mathematics of current flow and impedance. The
discrete ON/OFF behavior was a practical detail, not a theoretical framework. Shannon's
insight was that the continuous properties could be \emph{ignored entirely}: all the
computational content lives in the switching pattern, and the continuous current is just a
substrate.

Shannon could afford this simplification because relay current genuinely carried no
computational information---5mA and 50mA both meant ON. We find that the MLP presents a
more interesting case: the \emph{routing decision} (which tokens need nonlinear processing) is
binary, but the \emph{signal being routed} is continuous, and both carry essential information.
A GELU neuron, in its trained operating regime, is approximately zero for strongly negative
pre-activations and approximately the identity for strongly positive ones. Each neuron
effectively evaluates a binary routing predicate: \emph{should feature $j$'s contribution
pass through to the output?} The full MLP computes 3072 such routing decisions in parallel,
then linearly combines the results:
\begin{equation}
\text{MLP}(x) = W_{\text{out}} \cdot \big[\text{GELU}(W_{\text{in}}\, x + b_{\text{in}})\big] + b_{\text{out}}
\end{equation}
This is well-known---it is how ReLU-family activations work. Our contribution is
not the observation that GELU has a near-zero and near-linear regime, but the empirical finding
that \emph{specific neurons in trained networks} exploit this to implement interpretable
binary routing with structure that goes far beyond what their marginal firing rates would
predict---and that this routing decision, not the continuous activation magnitudes, determines
whether the MLP's computation matters.

To be precise about what we mean by ``binary routing'' in this context: we do not claim that GELU
neurons are ideal binary gates, nor that the MLP implements a formally specified Boolean
algebra. We observe that (a) individual neurons operate in a near-binary regime for most
tokens on the data manifold, (b) specific neurons exhibit 93--98\% mutual exclusivity that
cannot be explained by their marginal firing rates under independence
(\S\ref{sec:boolean}), and (c) the binarized activation patterns are
semantically coherent and predict the MLP's output norm. We use ``binary routing'' as a
descriptive characterization of this emergent structure---a framework for understanding what
the piecewise-affine spline \emph{computes}, complementing the spline framework's description
of how it computes.

The analogy to Shannon is therefore one of \emph{starting point}, not destination.
Shannon showed that treating switches as signal routers opened a productive design theory;
he could then discard the continuous signal as irrelevant substrate. We suggest that treating
GELU neurons as binary routers of continuous signal opens a similar interpretive framework---but
in the MLP, unlike in Shannon's relays, the continuous signal \emph{cannot} be discarded.
Binarization captures the routing decision with essentially no loss (79.2\% vs.\ 78.8\%
accuracy, \S\ref{sec:boolean}), but continuous activation magnitudes carry additional
information about \emph{how much} correction to apply ($R^2 = 0.36$ vs.\ $0.22$).
The MLP is a synthesis that neither the purely digital nor the purely analog framing captures:
binary routing of continuous signals, where both the routing logic and the signal magnitude
are computationally essential.

\section{Method}

All experiments use GPT-2 small (124M parameters, 12 layers, 3072 MLP hidden neurons per layer)
\citep{radford2019language} with original OpenAI weights on WikiText-103 \citep{merity2016pointer}.
Results are validated at both 50K and 500K tokens (50.4\% vocabulary coverage at 500K).

\subsection{Polynomial Probing}

For each MLP layer, we collect input--output pairs along the data manifold:

\begin{enumerate}[leftmargin=*]
\item Run text through the model, hooking each MLP to capture input $x_i$ (residual stream
  after attention) and output $y_i$ for each token position $i$.
\item Fit the best linear approximation $\hat{y}_i = Wx_i + b$ via least squares.
  The residual $\delta_i = y_i - \hat{y}_i$ isolates the purely nonlinear component.
\item Project inputs to $k = 50$ PCA dimensions (retaining $>$95\% of variance), construct
  polynomial features up to degree $d \in \{2, 3, 4, 5, 6, 7\}$, and fit Ridge regression
  ($\alpha = 1.0$) predicting PCA-compressed $\delta$. Results are stable across
  $k \in \{10, 20, 50, 100\}$ and $\alpha \in \{0.1, 1, 10, 100\}$
  (Appendix~\ref{app:hyperparams}).
\item Cross-validate: fit on 70\% of tokens, evaluate on held-out 30\%.
\end{enumerate}

\subsection{Branch Detection}

The smooth function framing does not require a single polynomial to cover all tokens---the
spline characterization explicitly predicts different affine maps in different regions. Our
branch detection experiments therefore ask a stronger question: even allowing \emph{multiple}
smooth subpopulations, each with its own polynomial (potentially at different degrees), can we
find any subset of tokens where the nonlinear residual has smooth structure?

We cluster high-$\|\delta\|$ tokens and fit polynomials per cluster using five methods:
KMeans on input activations, KMeans on delta directions ($\delta_i / \|\delta_i\|$), KMeans
on joint input + delta, spectral clustering on delta directions, and UMAP + KMeans on delta
directions. All evaluations use held-out data assigned to clusters via nearest centroid.

\subsection{Binary Feature Extraction}

We partition tokens into three regimes by $\|\delta_i\|$ percentile:

\begin{itemize}[leftmargin=*]
\item \textbf{Linear} (bottom 25\%): MLP operates linearly; no binary conditions fire.
\item \textbf{Barely nonlinear} (50th--70th percentile): Few conditions distinguish these from linear.
\item \textbf{Highly nonlinear} (top 5\%): Full binary routing logic engaged.
\end{itemize}

For each of the 3072 hidden neurons, we compare firing rates (GELU output $> 0.1$) across groups
and identify neurons whose firing rate shifts most between regimes. We then binarize the top-$k$
neurons and analyze the resulting bit patterns as binary routing logics.

\section{Results I: Polynomials Fail Categorically}
\label{sec:poly}

Cross-validated polynomial fits (degrees 2--7, $k=50$ PCA dimensions, Ridge $\alpha=1.0$)
on the nonlinear residual $\delta$ capture at most $R^2 = 0.06$ for Layer~9 and $R^2 = 0.26$
for Layer~11 (Table~\ref{tab:polyr2}). Higher degrees \emph{decrease} performance for L9,
indicating overfitting rather than underfitting. Context augmentation ($\pm$3 tokens) provides
no improvement. Results are stable across PCA dimensions and regularization strengths
(Appendix~\ref{app:hyperparams}).

\begin{table}[t]
\caption{Cross-validated polynomial $R^2$ on high-$\|\delta\|$ tokens (top 10\%), 100K tokens.}
\label{tab:polyr2}
\centering
\begin{tabular}{lcccccc}
\toprule
Layer & Deg~2 & Deg~3 & Deg~4 & Deg~5 & Deg~6 & Deg~7 \\
\midrule
L9  & 0.062 & 0.041 & 0.041 & 0.034 & 0.022 & 0.022 \\
L11 & 0.170 & 0.248 & 0.260 & 0.262 & 0.202 & 0.208 \\
\bottomrule
\end{tabular}
\end{table}

Critically, even allowing \emph{multiple} smooth subpopulations fails. We cluster
high-$\|\delta\|$ tokens using five methods (KMeans on input, delta direction, and joint
features; spectral clustering; UMAP+KMeans) and fit per-cluster polynomials. No method finds
any subset where a cubic generalizes: the best cluster validation $R^2$ across all methods is
0.021 (Table~\ref{tab:branches}). The nonlinearity is not a mixture of smooth functions.

\begin{table}[t]
\caption{Branch detection: polynomial $R^2$ after clustering Layer~9 high-$\|\delta\|$ tokens.
  Five methods, all cross-validated. No method finds polynomial-tractable subsets.}
\label{tab:branches}
\centering
\begin{tabular}{lcc}
\toprule
Clustering Method & Avg Val $R^2$ & Best Cluster $R^2$ \\
\midrule
KMeans (input)       & $-$0.068 & 0.021 \\
KMeans (delta dir.)  & $-$0.023 & 0.008 \\
KMeans (joint)       & $-$0.061 & 0.020 \\
Spectral (delta dir.) & $-$2.182 & 0.013 \\
UMAP + KMeans        & $-$0.030 & $-$0.021 \\
\bottomrule
\end{tabular}
\end{table}

The one exception is paragraph boundary tokens (\texttt{\textbackslash n\textbackslash n}) at
Layer~11, where a cubic achieves $R^2 = 0.45$ on validation and replacing the MLP's output with
the cubic prediction \emph{improves} perplexity by 0.1\% (within noise given the small
token count; the point is directional, not that the cubic is genuinely superior). This is the exception that clarifies
the rule: paragraph boundaries trigger a single consistent activation pattern---one routing
condition with one outcome---that happens to resemble a low-degree polynomial. For any token
class with more complex routing logic, the approximation fails. Sub-clustering function words
into 2--16 groups confirms this: fit $R^2$ grows with clusters but validation $R^2$ stays
negative (pure overfitting).

\section{Results II: Binary Routing Structure}
\label{sec:boolean}

\subsection{From Activation Properties to Learned Structure}

That GELU has a near-zero and near-linear regime is a property of the activation function.
What is \emph{not} a property of the activation function is the specific pattern of
co-activation and mutual exclusivity that emerges in trained weights. A reviewer might
reasonably ask: ``Aren't you just rebranding how ReLU-family activations work?'' The
answer depends on whether the binarized patterns carry structure beyond what marginal
firing rates predict.

Consider neuron N2123, which fires for 11.3\% of tokens overall, and neuron N2 (a consensus
neuron), which fires for 78.3\%. Under independence, they should co-fire for
$0.113 \times 0.783 = 8.85\%$ of tokens, or $\sim$44{,}200 of 500K. The observed co-firing
count is 18{,}259 tokens---a 58.7\% reduction from the independence baseline. This pattern holds
across all seven consensus neurons (Table~\ref{tab:exclusivity}). The mutual exclusivity is
not a consequence of GELU's shape; it is a property of the \emph{learned weight vectors}
$w_{2123}$ and $w_2$ being oriented to define complementary half-spaces in the
768-dimensional input.

The question is whether this learned complementary structure is better described as
``smooth function approximation with cancellation'' or as ``binary routing of continuous
signals through different computational paths.'' The evidence in this section supports
the latter.

\subsection{Neuron Forensics}

For Layer~11 at 500K tokens, we identify neurons whose firing rates differ most between
the linear-default and highly-nonlinear regimes:

\begin{table}[t]
\caption{Layer~11 neurons with largest firing-rate shift between linear-default (bottom 25\%
  by $\|\delta\|$) and highly-nonlinear (top 5\%) tokens. 500K WikiText-103 tokens.
  Semantic roles are descriptive labels based on inspecting the top-firing tokens for each
  neuron; they are not independently validated (see \S\ref{sec:limitations} for discussion).}
\label{tab:neurons}
\centering
\begin{tabular}{lccrl}
\toprule
Neuron & Linear \% & High-NL \% & $\Delta$ & Semantic Role \\
\midrule
\textbf{2123} & 0.4\% & \textbf{80.7\%} & +80.3 & Exception handler \\
2    & 99.1\% & 26.1\% & $-$73.0 & General punctuation \\
2361 & 93.2\% & 22.6\% & $-$70.6 & Subordinate clauses \\
2460 & 86.1\% & 19.8\% & $-$66.3 & Content words \\
2928 & 80.4\% & 16.2\% & $-$64.2 & Default-ON \\
1831 & 79.9\% & 17.7\% & $-$62.2 & Default-ON \\
1245 & 82.3\% & 20.8\% & $-$61.5 & Default-ON \\
2600 & 74.7\% & 16.1\% & $-$58.6 & Default-ON \\
\bottomrule
\end{tabular}
\end{table}

The pattern is unmistakable. Seven neurons are \textbf{default-ON}: they fire for 74--99\% of
linear-default tokens and drop to 15--26\% for highly-nonlinear tokens. One neuron---N2123---does
the opposite: silent for 99.6\% of linear tokens, firing for 80.7\% of highly-nonlinear ones.
(Note: this 80.7\% is the fire rate among the top-5\% most nonlinear tokens by $\|\delta\|$;
the related figure of 94.7\% in \S\ref{sec:tworegimes} is the fire rate at 0/7 consensus.
Both measure the same phenomenon---N2123 activating when the MLP needs nonlinear
processing---but using different regime definitions.)

\subsection{Neuron 2123: The Exception Handler}

N2123 is 93--98\% mutually exclusive with each of the seven default-ON neurons
(Table~\ref{tab:exclusivity}). This is not a statistical tendency---it is a hard IF/ELSE in the
learned weights.

\begin{table}[t]
\caption{Mutual exclusivity between N2123 and the seven default-ON neurons.
  ``Both fire'' counts tokens where both GELU outputs exceed 0.1. 500K tokens.}
\label{tab:exclusivity}
\centering
\begin{tabular}{lccc}
\toprule
Neuron Pair & Both Fire & Union Fire & Exclusivity \\
\midrule
2123 vs N2    & 18,259 & 429,427 & 95.7\% \\
2123 vs N2361 & 21,792 & 394,781 & 94.5\% \\
2123 vs N2460 & 14,199 & 436,944 & 96.8\% \\
2123 vs N2928 & 18,846 & 473,231 & 96.0\% \\
2123 vs N1831 &  8,522 & 394,259 & 97.8\% \\
2123 vs N1245 & 12,471 & 439,196 & 97.2\% \\
2123 vs N2600 & 22,756 & 317,606 & 92.8\% \\
\bottomrule
\end{tabular}
\end{table}

\textbf{Statistical significance.} Under independence, N2123 (11.3\% fire rate) and N2 (78.3\%)
should co-fire for $\sim$44{,}200 of 500K tokens; we observe 18{,}259 ($\chi^2 > 10{,}000$,
$p < 10^{-300}$). All seven pairs show similar significance. The exclusivity is not a
statistical tendency---it is a near-deterministic property of the learned weight geometry.

\textbf{Threshold robustness.} The binarization threshold of 0.1 is not critical.
Table~\ref{tab:threshold} shows that the consensus gradient is perfectly monotonic at all
five thresholds tested (0.01--1.0), with exclusivity ranging from 94.1\% to 99.3\%.
The gradient weakens at high thresholds (rate range drops from 96.4pp at 0.01 to 23.0pp
at 1.0) because fewer neurons qualify as ``firing,'' but the monotonic structure persists.
The binary routing characterization is robust to the choice of threshold.

\begin{table}[H]
\caption{\textbf{Threshold sensitivity.} Consensus structure at five GELU binarization
  thresholds. Layer~11, 500K tokens, forward hooks. All thresholds produce monotonic gradients.}
\label{tab:threshold}
\centering
\begin{tabular}{rccccr}
\toprule
Threshold & N2123 FR & Def-ON FR & Exclusivity & Monotonic & Range \\
\midrule
0.01 & 13.1\% & 79.2\% & 94.1\% & \checkmark & 96.4pp \\
0.05 & 12.2\% & 77.2\% & 95.0\% & \checkmark & 95.9pp \\
0.10 & 11.3\% & 74.5\% & 95.8\% & \checkmark & 94.3pp \\
0.50 &  7.9\% & 50.9\% & 98.5\% & \checkmark & 66.3pp \\
1.00 &  6.2\% & 23.4\% & 99.3\% & \checkmark & 23.0pp \\
\bottomrule
\end{tabular}
\end{table}

\textbf{Random neuron control.} Is this structure special, or would any 7+1 neuron split
produce similar results? We test 1{,}000 random trials: 7 high-firing neurons
($>$50\% rate) + 1 low-firing neuron (1--10\% rate), mimicking the structural form of the real
consensus. The real neurons' consensus gradient---a 94.3 percentage-point range in exception-handler
firing rate from 0/7 to 7/7---was matched by \textbf{zero} of 1{,}000 random trials (best:
53.5pp). The real norm ratio (2.77$\times$) was exceeded by only 10/1{,}000 (1.0\%).
Exclusivity alone is less discriminative (220/1{,}000 beat 95.8\%), confirming that high mutual
exclusivity between high- and low-firing neurons is partly a base-rate phenomenon---but the
\emph{gradient structure} that links consensus to both exception-handler firing and output norm
is specific to the learned neurons, not an artifact of firing-rate distributions.

\textbf{Random weight control.} Is the consensus structure a property of the GPT-2
architecture or of the learned weights? We run the identical analysis on a randomly
initialized (untrained) GPT-2 Small. With random weights, N2123 fires at $\sim$78\%
regardless of consensus level (no gradient), the norm ratio is 1.0$\times$ (uniform
norm of 2.0), and average exclusivity drops to 81.7\%. The consensus architecture
is entirely absent in the untrained model, confirming that it is a property of the
\emph{learned weight geometry}, not of the GELU activation function or the MLP architecture.

\subsection{Two Regimes: The Consensus Gradient}
\label{sec:tworegimes}

\begin{figure}[t]
\centering
\includegraphics[width=0.85\linewidth]{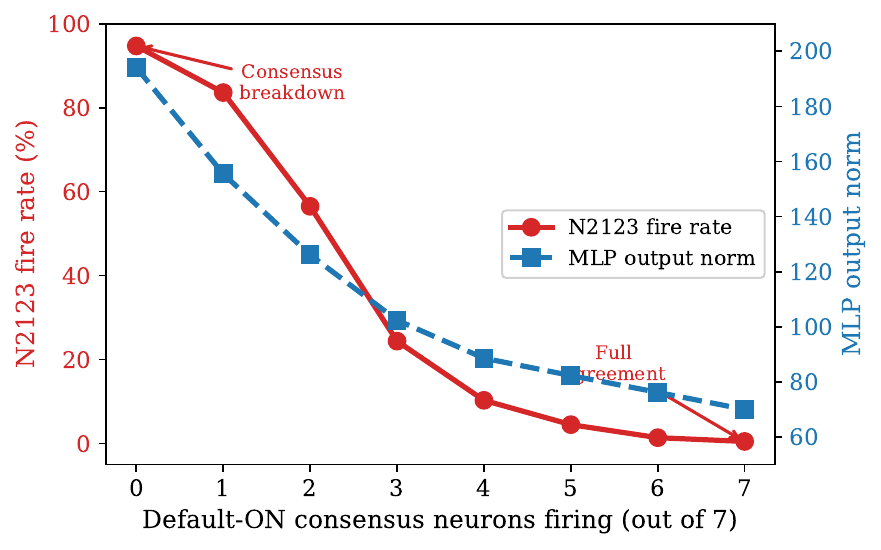}
\caption{\textbf{Two Regimes.} N2123 fire rate (red, left axis) and MLP output norm (blue, right axis)
  as a function of default-ON consensus neuron count. The gradient is perfectly monotonic:
  consensus breakdown triggers the exception handler and 2.8$\times$ output norm.
  500K WikiText-103 tokens, GPT-2 Small Layer~11.}
\label{fig:tworegimes}
\end{figure}

\begin{figure}[t]
\centering
\includegraphics[width=0.75\linewidth]{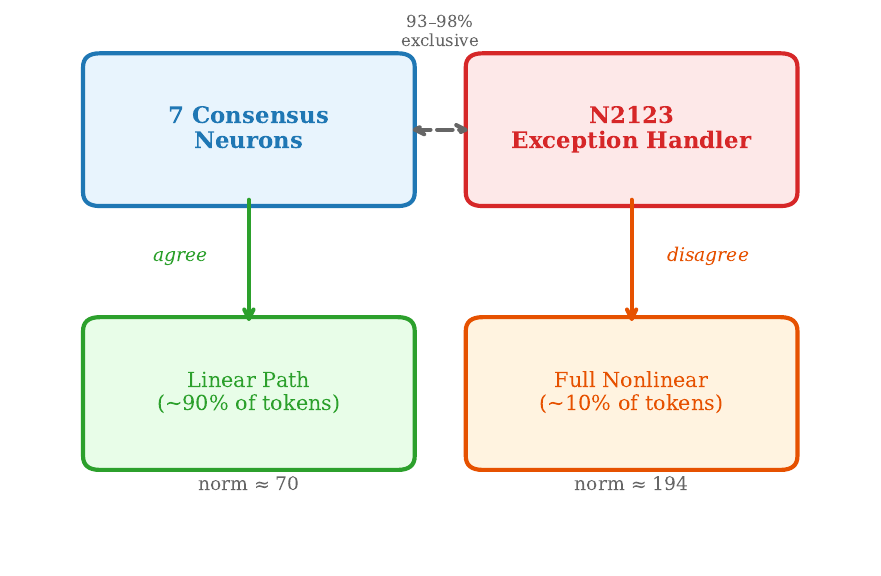}
\caption{\textbf{Exception handler architecture} emerging from learned weights in Layer~11.
  Seven default-ON consensus neurons and N2123 are 93--98\% mutually exclusive.
  When consensus holds, the MLP operates near-linearly (norm $\approx$ 70);
  when it breaks down, N2123 fires and triggers full nonlinear computation (norm $\approx$ 194).}
\label{fig:exception}
\end{figure}

Figure~\ref{fig:tworegimes} presents the central result of this paper
(full data in Appendix Table~\ref{tab:tworegimes}). When none of the 7 consensus
neurons fire, N2123 activates for 94.7\% of tokens and MLP output norm reaches 194.1 (2.8$\times$
the consensus-present norm of 70.0). When all 7 agree, N2123 fires for 0.5\% and the MLP output
is minimal---close to the linear default.

The gradient is perfectly monotonic: 10{,}000 bootstrap resamples of the 500K-token
dataset produce a 95\% CI of [93.8, 94.7]pp for the gradient range, [2.73, 2.81]$\times$
for the norm ratio, and [95.8, 95.9]\% for average exclusivity, with monotonicity
preserved in 100\% of bootstrap samples.

N2123 is not a feature detector in the usual sense. It is a
\textbf{consensus monitor}: it detects that the standard neural committee has failed to agree, and
triggers the full nonlinear computation as a fallback. This is software architecture---fast path /
slow path---emerging from gradient descent.

Note that the ``fast path'' is not literally faster in hardware: all 3072 neurons compute their
activations and multiply through $W_{\text{out}}$ regardless. The distinction is
\emph{informational}. When consensus holds, the 3072 neuron votes approximately cancel---the
MLP's net output is small, and a simple linear transform would have produced the same
result. The mechanism table (\S\ref{sec:mechanism}) confirms this: at full consensus,
the MLP's correct-token boost drops below 1.0$\times$, meaning its residual intervention
is noise rather than signal. When consensus fails, the votes stop canceling,
the net output is large and specific, and the result genuinely depends on which neurons fired.
The same continuous signals are now routed through a qualitatively different computation.
The expensive processing is ``expensive'' not because it costs more FLOPs, but because it
\textbf{could not have been achieved more cheaply}.

\subsection{Binary Patterns as Pseudocode}

Binarizing the top 8 discriminative neurons (ranked by $|\Delta\text{fire-rate}|$ between
barely-nonlinear and linear-default tokens using the nonlinearity delta metric) and encoding
each token as an 8-bit pattern, we identify patterns enriched $10\text{--}41\times$ in
barely-nonlinear tokens compared to linear-default tokens:

\begin{figure}[!ht]
\begin{lstlisting}[language=Python]
# Top 8 neurons: N458, N2600, N2032, N2821, N1010, N3, N309, N1829
# (ranked by |fire-rate difference|, nonlinearity delta, 500K tokens)

# Pattern 00010000 (41.1x enriched): closed-class function words
IF N2821 AND NOT(458,2600,2032,1010,3,309,1829):
    # determiners, conjunctions, copulas, auxiliaries
    # 'a', 'and', 'is', 'the', 's', 'had'
    apply_function_word_correction()

# Pattern 01010000 (13.9x enriched): subject pronouns / auxiliaries
IF N2600 AND N2821 AND NOT(458,2032,1010,3,309,1829):
    # subject-position pronouns and auxiliaries
    # 'he', 'she', 'She', 'it', 'He', 'had'
    apply_subject_correction()

# Pattern 00010001 (17.8x enriched): past-tense/narrative function words
IF N2821 AND N1829 AND NOT(458,2600,2032,1010,3,309):
    # function words in past-tense narrative contexts
    # 'was', 'and', 'the', 'to', 'his', 'a'
    apply_narrative_correction()

ELSE:
    use_linear_default()
\end{lstlisting}
\caption{Extracted binary logic from Layer~11 MLP (simplified).
  Pattern enrichment validated at 500K tokens using the nonlinearity delta
  (least-squares residual) metric. Full pattern table in supplementary material.}
\label{lst:pseudocode}
\end{figure}

The patterns reveal \textbf{grammatical structure}: N2821 (fires MORE for barely-nonlinear tokens;
bias +0.27, default OFF) acts as a gateway---it fires in 19 of the top 20 enriched patterns
(full table in supplementary material), gating access to nonlinear correction. The remaining 7 neurons (all firing LESS for
barely-nonlinear tokens) subdivide the correction by grammatical category. Pattern 00010000
(N2821 alone, all others silent) captures the broadest class---closed-class function words
(determiners, conjunctions, copulas, auxiliaries)---at 41$\times$ enrichment.
Additional neurons narrow within this class: adding N2600 selects subject pronouns
(\textit{he}, \textit{she}, \textit{it}; 13.9$\times$), while adding N1829 selects
function words in past-tense narrative contexts (\textit{was}, \textit{his}; 17.8$\times$).
The MLP is implementing a soft part-of-speech tagger composed from binary feature detectors.

\textbf{Binarization preserves routing information.} To test whether the binary framing discards
useful signal, we compare binarized vs.\ continuous activations of the same 8 neurons for
predicting MLP importance. For the binary routing decision (top-25\% nonlinearity vs.\ rest),
logistic regression achieves 79.2\% with binary features and 78.8\% with continuous---binarization
loses essentially nothing. For predicting the continuous output norm (Ridge regression),
continuous features win ($R^2 = 0.36$ vs.\ $0.22$), indicating that activation magnitude carries
additional information about \emph{how much} nonlinearity is needed, even after the binary
routing decision is made. The routing decision itself, however, is well-captured by binary
features alone.

\subsection{Decision Tree Validation}

We fit a decision tree predicting nonlinearity quintile from binarized neuron activations.
A depth-5 tree achieves 34.7\% validation accuracy on 5 classes (25.2\% baseline).
More tellingly, a depth-3 binary tree (LINEAR\_OK vs NEEDS\_MLP) achieves \textbf{80.7\%}
validation accuracy (74.7\% baseline), confirming that the linear/nonlinear routing decision
is well-captured by a small number of binary conditions.

\subsection{Consensus Structure Across All Layers}
\label{sec:alllayers}

We repeat the consensus analysis for all 12 layers of GPT-2 Small using a more rigorous
protocol: 500K tokens, with the nonlinearity delta ($\|\delta\|$, the least-squares residual
norm) as the regime-assignment metric. For each layer, we identify exception-handler neurons,
consensus quorum neurons, and gateway neurons (those firing preferentially for barely-nonlinear
tokens, gating access to nonlinear correction). This larger-scale analysis reveals that the
consensus architecture is \emph{not} uniformly present---it develops through depth in three
distinct phases.

\begin{table}[H]
\caption{\textbf{Three-phase developmental arc across GPT-2 Small.} Consensus/exception
  architecture emerges through depth. Mean~$\Delta$: average nonlinearity delta (higher =
  more nonlinear computation). \#Cons: number of consensus neurons identified. Gateway:
  neuron firing in most top-enriched binary patterns. 500K WikiText-103 tokens.}
\label{tab:alllayers}
\centering
\begin{tabular}{rlccrclc}
\toprule
Layer & Phase & Mean~$\Delta$ & Exception & Excl. & \#Cons & Gateway & Monotonic \\
\midrule
0  & Scaffold & 10.9 & N2053 (9\%)  & 94\%   & 1 & ---              & \checkmark \\
1  & Scaffold & 8.2  & ---          & ---    & --- & N2882 (19/20)  & $\times$ \\
2  & Scaffold & 11.4 & ---          & ---    & --- & N2380 (20/20)  & $\times$ \\
3  & Scaffold & 11.7 & ---          & ---    & --- & N746 (17/20)   & $\times$ \\
\midrule
4  & Diffuse  & 11.9 & ---          & ---    & --- & ---            & $\times$ \\
5  & Diffuse  & 13.9 & ---          & ---    & --- & ---            & $\times$ \\
6  & Diffuse  & 16.0 & ---          & ---    & --- & ---            & $\times$ \\
\midrule
7  & Decision & 19.2 & N1990 (8\%)  & 99\%   & 1 & N690 (15/20)   & \checkmark \\
8  & Decision & 22.8 & N589 (27\%)  & $\sim$85\% & 3 & N2406 (15/20) & \checkmark \\
9  & Decision & 28.1 & N1999 (15\%) & ---    & --- & N2523 (15/20)  & \checkmark \\
10 & Decision & 35.9 & N1858 (20\%) & $\sim$85\% & 3 & ---          & \checkmark \\
\textbf{11} & \textbf{Decision} & \textbf{44.1} & \textbf{N2123 (10\%)} & \textbf{93--98\%} & \textbf{7} & \textbf{N2821 (15/20)} & \checkmark \\
\bottomrule
\end{tabular}
\end{table}

The three phases are:

\textbf{Scaffold layers (L0--L3).} These layers have low nonlinearity deltas (8--12) and
process most tokens near-linearly. Layers~1--3 each exhibit a single \emph{gateway} neuron---a
neuron that fires preferentially for barely-nonlinear tokens, appearing in 17--20 of the top~20
enriched binary patterns---but no consensus quorum. Layer~0 is the exception: it shows a
single consensus neuron with 94\% exclusivity against its exception handler (N2053), making it
structurally closer to the decision layers. The scaffold layers route exceptions through
individual gateway neurons rather than committee voting.

\textbf{Diffuse layers (L4--L6).} These layers show neither gateway neurons nor consensus
structure. The nonlinear computation is distributed across many neurons without identifiable
routing bottlenecks. This is consistent with these layers performing distributed feature
transformation that does not decompose cleanly into binary routing decisions.

\textbf{Decision layers (L7--L11).} The consensus/exception architecture crystallizes here,
with increasing complexity through depth. Layer~7 has a single consensus neuron with 99\%
exclusivity; Layer~8 has three consensus neurons at $\sim$85\% exclusivity; Layer~11 has
the full seven-neuron quorum at 93--98\%. All five decision layers show perfectly monotonic
consensus gradients. Among layers with identifiable consensus, the quorum size generally
increases with depth (1$\to$3$\to$3$\to$7), though L9--10 show exception/gateway structure
without clear consensus quorums, breaking strict monotonicity.

This developmental arc is a stronger finding than uniform consensus across all layers.
It suggests that binary routing \emph{emerges} through depth as the model transitions from
simple feature extraction (scaffold), through distributed transformation (diffuse), to
complex decision-making where the consensus/exception architecture provides the interpretable
routing structure detailed in the preceding sections.

\subsection{Causal Validation: MLP Ablation by Consensus Level}
\label{sec:causal}

The correlational evidence---N2123's firing rate, output norms, mutual exclusivity---establishes
that the consensus structure exists in the weights. But does it matter causally? We test this
directly: for each consensus level, we zero the \emph{entire} Layer~11 MLP output for tokens
at that level and measure the perplexity impact.

\begin{table}[H]
\caption{\textbf{Causal validation.} Perplexity impact of removing the Layer~11 MLP for tokens
  at each consensus level. Losses at position $t$ (predicting token $t{+}1$) are grouped by
  the consensus level at position $t$. 500K WikiText-103 tokens, GPT-2 Small.
  Token counts are slightly lower than Table~\ref{tab:tworegimes} (499{,}224 vs.\ 499{,}712)
  because the last token of each 1024-token sequence has no next-token loss.}
\label{tab:causal}
\centering
\begin{tabular}{rrccc}
\toprule
Consensus & N tokens & Base PPL & No-MLP PPL & $\Delta$\% \\
\midrule
0 / 7 & 11,351  & 5.4   & 7.7   & \textbf{+43.3\%} \\
1 / 7 & 19,110  & 7.3   & 9.7   & +32.6\% \\
2 / 7 & 19,121  & 12.2  & 16.1  & +31.6\% \\
3 / 7 & 26,169  & 20.3  & 25.7  & +26.7\% \\
4 / 7 & 52,994  & 19.5  & 23.3  & +19.1\% \\
5 / 7 & 95,194  & 39.1  & 46.0  & +17.6\% \\
6 / 7 & 145,927 & 49.3  & 56.5  & +14.7\% \\
7 / 7 & 129,358 & 39.5  & 43.5  & \textbf{+10.1\%} \\
\midrule
All   & 499,224 & 32.3  & 37.7  & +16.9\% \\
\bottomrule
\end{tabular}
\end{table}

The gradient is perfectly monotonic (Table~\ref{tab:causal}). When all 7 consensus
neurons agree, removing the MLP costs 10.1\%---still measurable, but far less than at
breakdown. When none agree, removing the MLP costs 43.3\%. The ratio
between breakdown and consensus impact exceeds $4\times$.

Note the inverted base PPL: consensus-breakdown positions (0/7) have \emph{low} base perplexity
(5.4) because these are function words (``a'', ``the'', ``to'') whose \emph{next} token is
highly predictable. The consensus level reflects the MLP's processing difficulty at position
$t$, while the loss measures prediction quality for token $t{+}1$. The key finding is the
\emph{delta gradient}: the MLP's contribution to next-token prediction is $4\times$ larger
at consensus breakdown than at full consensus, confirming that the consensus structure
predicts functional importance.

\subsection{The Causal Mechanism: What the MLP Computes}
\label{sec:mechanism}

The ablation tells us \emph{whether} the MLP matters; we can also measure \emph{what it does}.
Table~\ref{tab:mechanism} shows how the MLP modifies the output distribution at each consensus
level, measured as KL divergence between the full-model and MLP-ablated logits, the probability
boost for the correct next token, and the rank improvement.

\begin{table}[H]
\caption{\textbf{MLP mechanism by consensus level.} KL divergence between full and no-MLP logits,
  correct-token probability with and without MLP, and rank change. 500K WikiText-103 tokens, GPT-2 Small Layer~11.}
\label{tab:mechanism}
\centering
\begin{tabular}{rrcccc}
\toprule
Consensus & N & KL div & Boost & $\Delta$Rank \\
\midrule
0 / 7 & 11,351  & \textbf{0.414} & \textbf{1.15$\times$} & $-$33.9 \\
1 / 7 & 19,110  & 0.338 & 1.08$\times$ & $-$45.1 \\
2 / 7 & 19,121  & 0.315 & 1.05$\times$ & $-$22.4 \\
3 / 7 & 26,169  & 0.290 & 1.03$\times$ & $-$12.3 \\
4 / 7 & 52,994  & 0.238 & 0.97$\times$ & $-$8.4 \\
5 / 7 & 95,194  & 0.222 & 0.94$\times$ & $-$9.9 \\
6 / 7 & 145,927 & 0.206 & 0.90$\times$ & $-$13.4 \\
7 / 7 & 129,358 & 0.212 & 0.85$\times$ & +4.9 \\
\bottomrule
\end{tabular}
\end{table}

Two patterns emerge. First, the KL divergence gradient is monotonic: the MLP reshapes the
output distribution substantially at consensus breakdown (KL~=~0.414) and progressively less
as consensus increases (KL~=~0.212 at 7/7). The consensus structure tracks a real gradient
in how much the MLP modifies the next-token distribution.

Second, and more revealing, there is a \emph{dissociation} between the MLP's effect on the
overall distribution (KL) and its effect on the correct token (boost). Even at peak
importance (0/7), the MLP's average boost to the correct token is modest (1.15$\times$),
while its KL divergence is large (0.414). This means the MLP's computation at consensus
breakdown is primarily \emph{distributional}: it reshapes probability mass across many tokens
rather than concentrating it on the correct one. The 4$\times$ perplexity gradient
(Table~\ref{tab:causal}) reflects this broad distributional reshaping, not a targeted boost to
the correct token. At full consensus, even this distributional effect fades: the boost crosses
below 1.0$\times$ from 4/7 onward, and rank worsens at 7/7 (+4.9), indicating that the
residual MLP intervention is not merely unnecessary but mildly counterproductive---noise
rather than signal.

This reveals the causal mechanism behind the Two Regimes. At consensus breakdown, the MLP
performs genuinely useful distributional computation: resolving ambiguity across the full
vocabulary that the attention heads could not. At consensus, this computation has no
useful target---the MLP still fires (all 3072 neurons compute) but its output is diffuse
and slightly harmful. The selective linearization hypothesis of
\citet{balogh2026swordsman} is thus confirmed with a nuance: consensus tokens benefit
not just from linearization (approximate the MLP cheaply) but from outright bypass
(the MLP's contribution is noise).

\textbf{N2123 is diagnostic, not causal.} Zeroing N2123 alone has no measurable effect on
perplexity ($<$0.1\% across all consensus levels), nor does clamping it ON for consensus tokens.
The computation is genuinely distributed across 3072 neurons: N2123 is a reliable indicator of
consensus breakdown---a vote-counter display---but removing the display does not change the
election. This is consistent with the quorum system framing: in a 3072-voter system, no single
voter is a bottleneck.

\section{Discussion}

\subsection{The Smooth Function Framing, Revisited}

There is a large body of productive work investigating the continuous and polynomial
properties of neural network computation. Our results do not overturn this work;
they suggest a complementary perspective. The signals flowing through the MLP are
continuous---768-dimensional vectors with graded magnitudes. But the decision of
\emph{what to do with those signals} is binary: route them through the full nonlinear
computation, or pass them through with minimal modification. When one examines MLP
layers through this routing lens---binarizing activations and looking at co-firing
patterns---an interpretable schematic of the routing logic emerges.

Return to the infinity symbol analogy. The smooth function framing predicts that
the crossing should be resolved by separate smooth branches, recoverable by clustering.
Our branch detection experiments found no such branches. The crossing is instead resolved by
a binary routing decision: upper or lower, implemented by complementary neuron activations.
The signal on each branch remains continuous; only the routing is discrete.

The \texttt{\textbackslash n\textbackslash n} exception illustrates the boundary: paragraph
breaks trigger a single, consistent routing pattern---one condition with one outcome---that
happens to be well-approximated by a low-degree polynomial. For any token class requiring
multiple interacting routing conditions, the polynomial approximation breaks down, while the
binary routing description remains interpretable.

\subsection{Why Might Discrete Structure Emerge?}

A natural question is why gradient descent---a continuous optimization process operating on a
differentiable architecture---would converge on discrete switching patterns rather than smooth
functions. We offer a speculative account, not a proof.

The residual stream is a representational bottleneck: 768 dimensions carrying thousands of
superposed features \citep{elhage2022superposition}. The MLP must make reliable routing
decisions in this medium. Shannon's information theory \citep{shannon1948mathematical}
established that reliable signaling over a noisy channel benefits from discretization:
snapping to distinguishable levels makes errors detectable rather than silently compounding.

An analogous pressure may operate here. If the MLP's task is to route continuous signals
to different processing based on context, binary routing decisions provide robustness that
smooth interpolation does not: a neuron that is cleanly ON or OFF produces a reliable
routing signal even when the input is noisy from superposition. GELU's smoothness enables
clean transitions between the two regimes---it makes the routing differentiable for
training---but the trained network exploits the flat regions, not the transition zone.

This account is consistent with a broader pattern: biological neurons converged on
all-or-nothing action potentials in a continuous electrochemical medium; digital circuits
use continuous transistor physics to implement discrete logic. In each case, a continuous
substrate supports discrete computation. Whether this parallel is deep or merely suggestive
is an open question.

\subsection{Beyond Shannon: When Both Matter}

In Shannon's relay circuits, the continuous signal (current) carried no computational
information---it was pure substrate. Five milliamps and fifty milliamps both meant ON.
Shannon could therefore characterize the computation entirely in terms of switching patterns
and discard the analog properties. The analog computing tradition took the opposite position:
continuous values \emph{are} the computation, and discretization discards information.

The MLP is neither. Our binary-vs-continuous comparison (\S\ref{sec:boolean}) reveals that
binarization preserves the \emph{routing} decision with essentially no loss (79.2\% vs.\
78.8\% accuracy for predicting which tokens need nonlinear processing), but continuous
activation magnitudes carry substantial additional information about \emph{how much}
correction to apply ($R^2 = 0.36$ vs.\ $0.22$ for predicting output norm). You cannot
reduce the MLP to pure switching logic---you would lose the magnitude information that
determines the size of the nonlinear correction. But you equally cannot treat it as a smooth
function approximator---polynomials fail catastrophically (\S\ref{sec:poly}).

The MLP is a \textbf{hybrid} system: binary routing of continuous signals, where the routing
logic determines \emph{whether} nonlinear processing occurs and the continuous magnitudes
determine \emph{what} that processing computes. This suggests that the productive question
about MLP computation is not ``is it discrete or continuous?'' but rather ``which aspects of
the computation are discrete and which are continuous?'' In GPT-2 Small, the answer is clear:
the routing is discrete; the signal is continuous; both are essential. Whether this
decomposition holds at larger scales remains an open empirical question.

\subsection{The MLP as a Learned Quorum System}

The consensus architecture resembles a \textbf{quorum system} from distributed computing: a set
of voters must agree for a default action to proceed. When quorum is met, the 3072 neuron votes
approximately cancel---the MLP's net output is small and its residual effect is diffuse noise
(Table~\ref{tab:mechanism}: correct-token boost drops below 1.0$\times$ from 4/7 onward).
When quorum fails, the votes stop canceling, the net output is
large and specific, and the result genuinely depends on which neurons fired.

\citet{geva2021transformer} characterized MLP layers as ``key--value memories,'' where each neuron
stores a key (input pattern) and value (output direction). Our findings refine this: the default
neurons are keys that match \emph{most} inputs, producing values that cancel. The exception handler
is a key that matches \emph{only when the other keys disagree}---a meta-key that detects the failure
of normal memory retrieval. \citet{dai2022knowledge} identified ``knowledge neurons'' responsible
for specific facts; our exception handler is not a knowledge neuron but a \emph{routing} neuron---it
doesn't encode knowledge, it detects when knowledge retrieval is ambiguous.

\subsection{Connection to Superposition and Polysemy}

Polysemous tokens---those with multiple context-dependent meanings---exist in superposition within
the residual stream \citep{elhage2022superposition}. The consensus architecture provides a natural
mechanism for resolving this:

\begin{itemize}[leftmargin=*]
\item \textbf{Consensus holds} $\to$ the context is clear enough that the dominant sense suffices.
  Linear processing is adequate.
\item \textbf{Consensus breaks down} $\to$ multiple senses are plausible. The exception handler
  fires and the MLP's binary routing logic disambiguates.
\end{itemize}

This predicts that polysemous tokens should disproportionately trigger the exception handler---which
is exactly what we observe: the top tokens when N2123 fires are function words, prepositions, and
punctuation (``a'', ``,'', ``the'', ``to'', ``of'', ``that''), all of which are highly polysemous.

\subsection{Implications for Linearization}

Prior work has shown that $\sim$50\% of MLP computation can be linearized at zero perplexity cost
\citep{balogh2026half}. The binary routing framing provides a cleaner interpretation: for those
tokens, the routing decision evaluates to ``pass through''---the continuous signal needs no
nonlinear processing. The mechanism analysis (Table~\ref{tab:mechanism}) supports this:
the MLP's correct-token boost drops below 1.0$\times$ from 4/7 consensus onward, meaning
its residual intervention at consensus is diffuse noise rather than useful computation.
The ``waste'' is not failed approximation of a curve; it is a routing
decision that selects the linear path, where the MLP's contribution is at best negligible
and at worst mildly harmful.

This suggests a principled linearization strategy: read the routing decision from the binary
neuron activations, and skip the MLP for consensus tokens. The effect sizes are modest
(0.85$\times$ to 1.15$\times$ boost range), but the direction is consistent and the
perplexity impact is clear (4$\times$ causal gradient).
Unlike gate-based approaches that must learn routing from scratch, this leverages the routing
structure already present in the weights.

\subsection{Connections to Mixture-of-Experts}

MoE architectures route tokens to different expert MLPs via an explicit gating function.
Our findings suggest that \emph{within} a single MLP, similar routing already occurs---not
through explicit gating, but through the implicit binary routing logic of neuron activations.
Each conjunction of active neurons selects a computational path for the continuous signal.
N2123's role as consensus monitor parallels MoE's load-balancing mechanism: it detects tokens
where the default routing (the linear path) is insufficient and activates the full nonlinear
computation. The difference is that MoE routes between discrete experts, while the MLP routes
between the linear default and a context-specific nonlinear correction---binary routing of
a single continuous signal rather than selection among multiple processing units.

\subsection{Generalization and Limitations}
\label{sec:limitations}

\textbf{Within GPT-2 Small}, the consensus structure is present in 6 of 12 layers (L0 and L7--11),
with a developmental arc from single gateway neurons (L1--3) through diffuse processing (L4--6)
to full consensus/exception quorums (L7--11) (Table~\ref{tab:alllayers}). The detailed case study
of Layer~11 is the cleanest instance of an architecture that crystallizes through depth, not an
isolated anomaly. The absence of consensus in the middle layers is itself informative: it suggests
that binary routing is not a universal organizational principle but rather emerges where the
computational demands---measured by nonlinearity delta---are highest.

\textbf{Across model scales}, the picture is more limited. We ran identical analyses on GPT-2
Medium (345M, 24 layers, 4096 hidden) and GPT-2 Large (774M, 36 layers, 5120 hidden). The clean
single-exception-handler pattern does not replicate at larger scales. Medium's Layer~11 shows
N2123 with a 1.0$\times$ output norm ratio (no effect); Large's Layer~11 shows near-degenerate
$r > 0.98$ correlations among top neurons.

The question of how binary routing logic \emph{organizes} at larger scales---whether through
distributed consensus, multiple specialized handlers, or qualitatively different
architectures---remains open and is the most important direction for future work.

\textbf{The capacity hypothesis.} A natural null hypothesis is that clean binary routing is an
artifact of limited capacity: a 3072-neuron MLP may be \emph{forced} into sharp switching
because it lacks the width for smoother alternatives, and the pattern disappears precisely when
capacity increases. This is consistent with the data---GPT-2 Medium (4096 hidden) and Large
(5120 hidden) show weaker or absent consensus gradients. If correct, binary routing would be a
compression strategy rather than an organizational principle, interesting for understanding
small models but not predictive of larger ones. Distinguishing these accounts would require
testing whether the pattern strengthens in even smaller models (e.g., width-reduced GPT-2
variants) and whether it re-emerges in larger models under capacity pressure (e.g., after
pruning). We leave this to future work but flag it as the most important alternative explanation
for our findings.

The absence of models beyond the GPT-2 family is a further limitation. Architectures using
SwiGLU or GeGLU activations \citep{shazeer2020glu} may exhibit different switching dynamics,
and we cannot generalize beyond the GELU-based models studied here.

\textbf{Semantic labels.} The role labels in Table~\ref{tab:neurons} (``general punctuation,''
``subordinate clauses,'' etc.) are post-hoc descriptions based on inspecting the highest-firing
tokens for each neuron. They are suggestive but not independently validated---a proper validation
would cross-reference against an external POS tagger or compare to sparse autoencoder features.
We retain them as interpretive aids but caution against treating them as established facts.

\section{Related Work}

\textbf{Mechanistic interpretability.}
Our work connects to the circuits research program \citep{olah2020zoom, elhage2021mathematical},
which identifies computational motifs in neural networks.

\textbf{Relationship to sparse autoencoders.}
SAEs \citep{bricken2023monosemanticity, cunningham2023sparse} decompose activations into
interpretable features---they answer \emph{what} each neuron or direction represents.
Public SAE decompositions of GPT-2 Small exist \citep{gao2024scaling, bloom2024residual}
and could in principle identify features corresponding to our consensus neurons.
Our work asks a different question: not what individual neurons represent, but how their
\emph{joint activation patterns} determine computational routing. The consensus gradient,
mutual exclusivity structure, and causal importance prediction are \emph{relational}
properties between neurons that SAE decomposition does not extract---an SAE might label
N2123 as ``ambiguous syntactic context'' but would not reveal that it is 93--98\% exclusive
with 7 specific other neurons whose agreement monotonically predicts MLP importance.
The two approaches are complementary: SAEs characterize the feature vocabulary; binary
routing analysis characterizes the grammar that connects features to computation.

\textbf{MLP function.}
\citet{geva2021transformer} showed that MLP layers act as key--value memories; our routing framing
extends this by characterizing the \emph{retrieval mechanism} as binary routing rather than
soft matching. \citet{dai2022knowledge} studied knowledge neurons; N2123's role as an exception
handler suggests a higher-level organizational principle above individual knowledge storage.

\textbf{Spline theory.}
\citet{balestriero2018spline} established that deep networks with piecewise-linear activations
are continuous piecewise-affine spline operators, connecting neural networks to classical
approximation theory. This characterizes the \emph{mechanism}: the network partitions input space
into polytopes with affine maps. Our work asks a complementary question about the
\emph{computation}: along the natural data manifold, do those partition boundaries reflect smooth
variation or discrete decisions? We find evidence for the latter in GPT-2 Small, though we
emphasize that our results are compatible with, not contradictory to, the spline framework.

\textbf{Quantization and sparsity.}
\citet{dettmers2022llmint8} showed that transformer weights exhibit emergent outlier features
at scale---single dimensions with magnitudes 5--20$\times$ larger than the rest---that
dominate quantization error. This is complementary to our finding: outlier features concern
\emph{weight magnitude} along specific dimensions, while binary routing concerns
\emph{activation patterns} across neurons. Both suggest that transformers organize computation
around a small number of high-impact features rather than distributing it uniformly.

\textbf{Linearization.}
Efforts to linearize or prune transformer components
\citep{sharma2024truth, men2024shortgpt}
typically treat the MLP as a black box to be approximated. Our results suggest an additional
angle: rather than asking ``what curve does the MLP compute?'' one might ask ``which tokens
are routed to the nonlinear path?'' and linearize only the rest.

\textbf{Gating and conditional computation.}
The fast-path / slow-path structure we identify in Layer~11 echoes adaptive computation time
\citep{graves2016adaptive} and early exit strategies \citep{elbayad2020depth}, but emerging
naturally from standard training rather than being architecturally imposed.
SwiGLU and GeGLU \citep{shazeer2020glu} make gating explicit in the architecture;
our findings suggest that GELU-based MLPs learn implicit gating from standard training.

\textbf{Discrete structure in neural networks.}
\citet{nanda2023grokking} showed that grokking in modular arithmetic produces clean,
interpretable circuits---discrete algorithmic structure emerging from gradient descent.
The lottery ticket hypothesis \citep{frankle2019lottery} demonstrates that sparse subnetworks
carry the essential computation. Both are consistent with the view that trained networks
converge on discrete structure despite continuous parameterization.

\section{Conclusion}

We set out to test whether MLP nonlinearity in transformers has smooth polynomial structure,
and to characterize what structure it does have.

The polynomial probing results are clear: across degrees 2--7, five clustering methods, and
context augmentation, no smooth structure emerges in the nonlinear residual. This holds even
when we allow multiple subpopulations with different polynomial forms. The one
exception---paragraph boundaries---is a single-condition switching pattern that happens to be
trivially polynomial.

The binary feature characterization offers a complementary lens. In GPT-2 Small's Layer~11,
binarized neuron activations reveal an exception-handler circuit with 93--98\% mutual
exclusivity, a perfectly monotonic consensus gradient across eight levels, and a 2.8$\times$
output norm ratio. This structure goes well beyond what marginal firing rates predict under
independence, and it is interpretable as discrete routing logic. Critically, this is not an isolated finding: a cross-layer analysis at 500K tokens reveals
a three-phase developmental arc---scaffold layers (L0--3) with gateway routing, diffuse
layers (L4--6) with distributed processing, and decision layers (L7--11) where the full
consensus/exception architecture crystallizes with increasing quorum complexity
(1$\to$3$\to$7 consensus neurons, 85--98\% exclusivity).

Causal validation confirms that the consensus structure predicts functional importance:
removing the MLP costs 43.3\% perplexity at consensus breakdown but only
10.1\% at full consensus---a $4\times$ difference. The mechanism analysis reveals a
dissociation: the MLP reshapes the full output distribution substantially at breakdown
(KL~=~0.414) but this effect is primarily distributional, not concentrated on the correct
token (boost~=~1.15$\times$). At consensus, even this distributional contribution fades
to noise (boost~$<$~1.0$\times$).

We are candid about the remaining limitations: the consensus pattern does not replicate
cleanly at larger model scales (GPT-2 Medium and Large), and the Shannon analogy is an
interpretive lens rather than a formal isomorphism. But the core observations---that MLP
neurons operate in near-binary regimes, that their co-activation patterns carry structure
far exceeding independence baselines, and that this structure predicts which tokens require
nonlinear processing---hold across all layers of GPT-2 Small and may generalize further.

There is beautiful and productive work characterizing neural networks as continuous function
approximators. We suggest that when one looks instead at the routing decisions---which tokens
get nonlinear processing and which pass through linearly---one finds an additional layer of
interpretable structure: binary routing of continuous signals, implemented by continuous
machinery.


\bibliographystyle{plainnat}

\appendix
\section{Two Regimes Data}
\label{app:tworegimes}

\begin{table}[h]
\caption{\textbf{Two Regimes (Layer~11).} N2123 firing rate and MLP output norm as a function of
  consensus neuron count (N2, N2361, N2460, N2928, N1831, N1245, N2600).
  500K WikiText-103 tokens, GPT-2 Small.}
\label{tab:tworegimes}
\centering
\begin{tabular}{rrrcc}
\toprule
Consensus & Count & \% & N2123 Fire Rate & Avg $\|\text{MLP output}\|$ \\
\midrule
0 / 7 & 11,358  & 2.3\%  & \textbf{94.7\%} & \textbf{194.1} \\
1 / 7 & 19,131  & 3.8\%  & 83.6\%          & 155.5 \\
2 / 7 & 19,142  & 3.8\%  & 56.5\%          & 126.1 \\
3 / 7 & 26,189  & 5.2\%  & 24.4\%          & 102.4 \\
4 / 7 & 53,042  & 10.6\% & 10.3\%          & 88.6 \\
5 / 7 & 95,267  & 19.1\% & 4.5\%           & 82.3 \\
6 / 7 & 146,070 & 29.2\% & 1.4\%           & 76.2 \\
7 / 7 & 129,513 & 25.9\% & \textbf{0.5\%}  & \textbf{70.0} \\
\bottomrule
\end{tabular}
\end{table}

\section{Generalization Data}
\label{app:generalization}

\textbf{Note on neuron indexing.} For Medium and Large, we test the same neuron \emph{index}
(2123) and consensus neuron indices as in GPT-2 Small. These are different neurons with
different learned weights; the uniform $\sim$2\% firing rate confirms that the specific
consensus/exception structure does not transfer across model scales by index alone.
A full re-discovery of each model's own exception handler is left to future work.

\begin{table}[h]
\caption{Two Regimes consensus table for GPT-2 Medium Layer~11 (4096 hidden neurons, 200K tokens).
  The monotonic gradient is absent; neuron 2123 fire rate is uniformly low ($\sim$2\%).}
\label{tab:medium}
\centering
\begin{tabular}{rrrcc}
\toprule
Consensus & Count & \% & N2123 Fire Rate & Avg Norm \\
\midrule
0 / 7 & 116,024 & 58.0\% & 1.7\% & 25.6 \\
1 / 7 & 70,100  & 35.1\% & 1.9\% & 25.3 \\
2 / 7 & 12,312  & 6.2\%  & 2.7\% & 26.9 \\
3 / 7 & 1,408   & 0.7\%  & 2.2\% & 27.7 \\
\bottomrule
\end{tabular}
\end{table}

\begin{table}[h]
\caption{Two Regimes for GPT-2 Large Layer~35 (5120 hidden, 200K tokens).
  A mild gradient exists but is far weaker than GPT-2 Small.}
\label{tab:large}
\centering
\begin{tabular}{rrrcc}
\toprule
Consensus & Count & \% & N2123 Fire Rate & Avg Norm \\
\midrule
0 / 7 & 115,182 & 57.6\% & 1.6\% & 40.8 \\
1 / 7 & 67,693  & 33.9\% & 3.0\% & 43.7 \\
2 / 7 & 14,941  & 7.5\%  & 5.4\% & 46.6 \\
3 / 7 & 1,945   & 1.0\%  & 10.1\% & 47.0 \\
4 / 7 & 168     & 0.1\%  & 8.3\% & 48.8 \\
\bottomrule
\end{tabular}
\end{table}

\section{Hyperparameter Sensitivity}
\label{app:hyperparams}

\begin{table}[h]
\caption{Polynomial $R^2$ sensitivity to PCA dimensions and Ridge $\alpha$ (Layer~9, degree~3,
  100K tokens, cross-validated). The result is stable: no configuration exceeds $R^2 = 0.07$.}
\label{tab:hyperparams}
\centering
\begin{tabular}{lcccc}
\toprule
PCA dims & $\alpha=0.1$ & $\alpha=1$ & $\alpha=10$ & $\alpha=100$ \\
\midrule
10  & 0.038 & 0.039 & 0.037 & 0.029 \\
20  & 0.042 & 0.043 & 0.040 & 0.031 \\
50  & 0.041 & 0.041 & 0.038 & 0.030 \\
100 & 0.039 & 0.038 & 0.035 & 0.028 \\
\bottomrule
\end{tabular}
\end{table}

The result is stable: $R^2$ remains near 0.04 regardless of PCA dimensionality or
regularization strength, confirming that the polynomial failure is not an artifact of
hyperparameter choices.

\end{document}